\pdfoutput=1

\documentclass[11pt]{article}

\usepackage[]{EMNLP2023}

\usepackage{times}
\usepackage{latexsym}
\usepackage[T1]{fontenc}

\usepackage[utf8]{inputenc}

\usepackage{microtype}

\usepackage{inconsolata}

\usepackage[figuresleft]{rotating}
\usepackage{amssymb}

\title{Probing LLMs for Joint Encoding of Linguistic Categories}

\author{
\textbf{Giulio Starace$^{\bigstar,1}$}, 
\textbf{Konstantinos Papakostas$^{1}$}, 
\textbf{Rochelle Choenni$^1$},\\
\textbf{Apostolos Panagiotopoulos$^1$},
\textbf{Matteo Rosati$^1$},
\textbf{Alina Leidinger$^1$ and}
\textbf{Ekaterina Shutova$^1$}
\\
\normalsize{$^1$Institute for Logic, Language and Computation, University of Amsterdam}}

\usepackage{graphicx}
\graphicspath{ {./images/} }
\usepackage{comment}
\usepackage{caption}
\usepackage{subcaption}

\usepackage{booktabs}
\usepackage{multirow}
\usepackage{multicol}

\newcommand{\cameraready}[1]{{{#1}}}

\usepackage{lipsum}
\usepackage{afterpage}

\newcommand\blfootnote[1]{%
  \begingroup
  \renewcommand\thefootnote{}\footnote{#1}%
  \addtocounter{footnote}{-1}%
  \endgroup
}

\begin{document}
\raggedbottom
\maketitle
\blfootnote{$^\bigstar$ Corresponding author: \href{mailto:giulio.starace@gmail.com}{giulio.starace@gmail.com}.}
\begin{abstract}
	Large Language Models (LLMs) exhibit impressive performance on a range of NLP tasks, due to the
	general-purpose linguistic knowledge acquired during pretraining. Existing model interpretability
	research~\citep{tenney_bert_2019} suggests that a linguistic hierarchy emerges in the LLM layers,
	with lower layers better suited to solving syntactic tasks and higher layers employed for semantic
	processing. Yet, little is known about how encodings of different linguistic phenomena interact
	within the models and to what extent processing of linguistically-related categories relies on the
	same, shared model representations. In this paper, we propose a framework for testing the joint
	encoding of linguistic categories in LLMs. Focusing on syntax, we find evidence of joint encoding
	both at the same (related part-of-speech (POS) classes) and different (POS classes and related
	syntactic dependency relations) levels of linguistic hierarchy. Our cross-lingual experiments show
	that the same patterns hold across languages in multilingual LLMs.
\end{abstract}

\section{Introduction}

\begin{figure*}[!t]
	\centering
	\includegraphics[width=\textwidth]{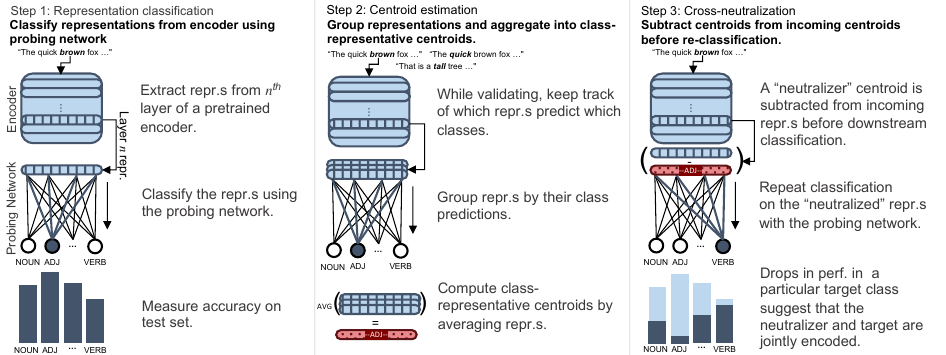}
	\caption{A diagram illustrating the three steps of our method: 1) Representation classification. 2)
		Centroid estimation. 3) Cross-neutralization. The complete methodology is outlined in
		Section~\ref{sec:method}.}
	\label{fig:method_overview}
\end{figure*}

Recent advancements in natural language processing (NLP) can be attributed to the development and
pretraining of large language models (LLMs) such as BERT, GPT-3, and many
others~\citep{devlin_bert_2019, brown_language_2020, touvron_llama_2023}. For their intended use of
providing general-purpose language representations suitable for many NLP tasks, these models must
efficiently capture a wide range of linguistic features within their finite capacity. Despite their
success, little is known about the way in which different types of linguistic information are
organized in these models. Systematically understanding how these models represent linguistic
phenomena and their interaction is crucial for the development of more effective NLP methods.

Existing research probed LLMs for their encoding of various linguistic properties such as
agreement~\citep{jawahar2019does}, word order and sentence
structure~\citep{tenney_what_2018,hewitt_structural_2019}, co-reference~\citep{tenney_bert_2019},
semantics~\citep{ettinger_what_2020} and multilinguality~\citep{ravishankar_multilingual_2019,
	libovicky_language_2020}. Taking a step further, \citet{tenney_bert_2019} and
\citet{clark_what_2019} studied where linguistic information is encoded in LLMs by probing different
layers. Their results demonstrated that a linguistic hierarchy emerges in BERT representations, with
lower layers capturing local syntax and higher layers being employed in higher-level semantic and
discourse tasks. However, we do not yet understand how encodings of different linguistic phenomena
interact within the models and to what extent processing of linguistically-related categories relies
on the same, shared model representations.

There are many dependencies between processing different linguistic phenomena: for instance,
information about a word’s part of speech is likely to be employed when disambiguating its word
sense. Alternatively, lower-level syntax is an important first step for semantic composition and
natural language understanding tasks. In this work, we investigate how the (hierarchical)
dependencies between different linguistic categories are encoded in LLMs, focusing on syntax. We ask
a set of novel questions: (1) how related syntactic categories (e.g. different parts of speech
(POS), such as \textit{noun} or \textit{verb}) are encoded within the models; (2) how syntactic
categories at different levels of the linguistic hierarchy (e.g. POS classes and syntactic
dependency relations) interact within the model; and (3) whether the observed patterns hold across
languages.

To answer these questions, we propose a framework for testing the joint encoding of linguistic
categories in LLMs. Specifically, we investigate to what extent the information about distinct
linguistic categories is shared in the parameters of the model. Our approach (see Fig.
\ref{fig:method_overview}) is inspired by the work of \citet{choenni_investigating_2022}, who
studied how LLMs share information across languages. We employ their \emph{cross-neutralization}
method, but extend it to study how information is encoded across linguistic categories in two
syntactic tasks---part-of-speech (POS) tagging and dependency (DEP) labeling. In short, we test
whether removing information on one syntactic category results in a failure to process another,
related one. For instance, would removing a representation of the \textit{verb} class hurt the
model's ability to identify the verbs' syntactic dependencies, while still succeeding in this task
on other categories, e.g.\ \textit{nouns}? This provides insight into whether these categories are
jointly encoded.

We study joint encoding patterns within both a monolingual and multilingual LLM, namely
RoBERTa~\citep{liu_roberta_2019} and XLM-R~\citep{conneau_unsupervised_2020}. For the latter, we
focus our analysis on English, Greek, and Italian, investigating to what extent multilingual models
encode information for POS tags and DEP relations in a language-agnostic manner. We find that POS
tags that are linguistically related are indeed jointly encoded by both the monolingual and
multilingual models, and observe similar joint encoding patterns across all three languages.
Moreover, we obtain further evidence of both language-agnostic and language-specific encoding within
multilingual models, given that representations specific to POS tags and DEP relations can be
approximately transferred across languages without a substantial impact on performance. Lastly, we
find evidence of joint encoding between related POS tags and DEP relations, suggesting information
sharing across tasks at different levels of the linguistic hierarchy.

\section{Background and related work}

A wide range of methods have emerged to study the inner workings of neural networks
\citep{belinkov_analysis_2019, madsen_post-hoc_2022}. Our approach is situated within the field of
probing~\citep{conneau_what_2018}, which typically involves the use of a simple auxiliary
``probing'' network \cameraready{trained to perform a specific task on the representations from a}
pretrained LLM. By keeping the probing network shallow and the pretrained model frozen, the
predictions can be used to identify which information was already captured by the pretrained LLM.

\paragraph{Previous work on probing}  \Citet{shi_does_2016} are the first to introduce the idea of
probing, training a logistic regression classifier on top of the embeddings from two neural machine
translation encoders to study the syntactic information learned by them. Similarly,
\citet{adi_fine-grained_2017} probe sentence representations for sentence length and word order by
training auxiliary task classifiers on predicting these exact properties from the representations.
Continuing on this work, \citet{conneau_what_2018} introduce a suite of probing tasks.
\Citet{tenney_bert_2019} apply these to BERT's hidden states, quantifying where linguistic
information is captured within the network. They find that the model follows the hierarchy of the
traditional NLP pipeline, with lower-level, syntactic features appearing earlier than more complex
semantic roles and structures. \citet{ravishankar_multilingual_2019} bring the same analysis to the
multilingual setting. \cameraready{\citet{dalvi_discovering_2021} find more supporting evidence of
	this through an unsupervised approach.} \Citet{manning_emergent_2020} probe the attention mechanism
in BERT for correspondence between linguistic phenomena such as syntactic dependencies and
coreference. They also make use of structural probes~\citep{hewitt_structural_2019}, finding that
the representations from BERT capture parse tree structure.

\paragraph{Syntactic knowledge in BERT} Various approaches have been used to study what knowledge
LLMs capture about syntax~\citep{rogers_primer_2021}. Previous probing studies have shown that BERT
embeddings encode information about parts of speech, syntactic chunks, constituent and dependency
labeling~\citep{tenney_what_2018, liu2019linguistic} as well as a broader set of syntactic features,
such as tree depth and tense~\citep{conneau_what_2018}. Yet, while it has become evident that syntax
is captured to some extent, less is still known about where and how this information is acquired.
Some works suggest that syntactic information is encoded in the token representations as they can be
used to successfully reconstruct syntactic trees~\citep{vilares_parsing_2020, kim_are_2019,
	hewitt_structural_2019}. Others have instead studied syntactic knowledge at the level of attention
heads, and show that particular heads specialize to specific aspects of syntax
\citep{htut_attention_2019, clark_what_2019}. However, \citet{htut_attention_2019} find that these
heads can not recover syntactic trees, suggesting that attention heads do not reflect the full
extent of syntactic knowledge that these models learn.

\paragraph{Information sharing in LLMs} \cameraready{Multiple works study information sharing in
	LLMs~\citep{blevins_deep_2018, sahin_linspector_2020}, with most focusing on cross-lingual
	sharing~\citep{chi_finding_2020,shapiro-etal-2021-multilabel-approach,stanczak-etal-2022-neurons}}.
\Citet{libovicky_language_2020} propose a simple method that removes language-specific information
from model representations by capturing it through the mean of a set of representations from the
respective language. \cameraready{\citet{ravfogel-etal-2020-null} iteratively remove gender
	information in word embeddings through projection onto the nullspace (INLP) in order to mitigate
	bias in biography classification~\citep{de-arteaga2019}. \Citet{elazar-etal-2021-amnesic} use INLP
	to build counterfactual representations for ``amnesic probing''. Here, the utility of a property for
	a given task is estimated by measuring the influence of removing the property via INLP, treating the
	removal as a causal intervention.} Our work is inspired by \citet{choenni_investigating_2022}, who
probe for joint encoding of typological features across different languages. In particular, they
probe LLMs for typological language properties and test whether subtracting language ``centroids''
from model representations negatively affects performance in typologically-related languages. While
they focus on representations of specific languages, we target the representations of linguistic
categories (e.g. nouns), and test whether these are jointly encoded across classes and tasks.

\paragraph{POS tagging and dependency labeling} We study the joint encoding of POS categories and
syntactic dependencies. Given a sentence, POS tagging is the task of mapping each word to the
appropriate part of speech. For instance, in ``The sailor dogs the hatch.'', ``dogs'' is a verb,
while in ``He chases the dogs'', ``dogs'' is a noun. Dependency labeling is the higher-level task of
labeling the dependency relation between a ``head'' word and a ``dependent'' (or ``parent'' and
``child''). For instance, in ``That is a black car.'', ``black'' is an \textit{adjectival modifier}
of ``car''.

\begin{table}[!t]
	\centering
	\footnotesize
	\begin{tabular}{@{}lcccc@{}}
		\toprule
		\multirow{2}{*}{} & \textbf{RoBERTa} & \multicolumn{3}{c}{\textbf{XLM-R}}                                       \\ \cmidrule(lr){2-2} \cmidrule(lr){3-5}
		                  & \textbf{en\_gum} & \textbf{en\_gum}                   & \textbf{it\_vit} & \textbf{el\_gdt} \\
		\textbf{POS}      & 95.6\%           & 95.5\%                             & 97.4\%           & 97.9\%           \\
		\textbf{DEP}      & 90.9\%           & 91.4\%                             & 93.9\%           & 94.8\%           \\ \bottomrule
	\end{tabular}
	\caption{Classification accuracy of our probing classifiers on English, Italian and Greek datasets
		for part-of-speech tagging and dependency labeling.}
	\label{tab:baseline_acc}
\end{table}

\section{Methodology}\label{sec:method}

\subsection{Probing}

To study the representations from a given encoder network, we \cameraready{train} a shallow classifier, or
``probing classifier" on a probing task, \cameraready{using the representations from the encoder as input.
	By keeping the encoder frozen}, we can ascribe all the learning to the relatively inexpressive probing
classifier, allowing us to probe whether the representations from the encoder contain the information
necessary to solve the task at hand.

\subsubsection{Probing tasks}\label{sec:probing_tasks}

We focus on POS tagging and DEP labeling as our probing tasks. For both cases, our encoders take
sentences as input, producing token-level embeddings at each layer. We extract the token-level
embeddings from a given layer and pool them into word embeddings (detailed in Section
\ref{sec:probe_config}). For POS tagging, the probing classifier receives the word embeddings and is
trained to classify them across $17$ categories. For DEP labeling, we aim to label the dependency
between child and parent. Thus, we pair each word in the sentence with its head as labeled in the
dataset and concatenate their representations\footnote{Different methods such as adding the mean
	vector or absolute difference of the pair resulted in similar performance.}. The probing classifier
takes this concatenation as input and is trained to classify the dependency relation between the
corresponding words across $36$ categories.

\subsubsection{Datasets}

We use the Universal Dependencies treebanks ~\citep{nivre_universal_2020}, manually annotated for
POS tagging and DEP labeling. We choose the GUM ~\citep{zeldes_gum_2017},
VIT~\citep{delmonte_vit_2017}, and the GDT~\citep{prokopidis_universal_2017} datasets for English,
Italian, and Greek respectively. All datasets contain word-level annotations with a total of 17 POS
tags and 36 DEP relations shared across languages\footnote{We summarize the split sizes and an
	overview of the pre-processing pipeline for all three languages in
	Appendix~\ref{app:dataset_preprocess}.}.

\begin{table}[!t]
	\footnotesize
	\centering
	\begin{tabular}{@{}lcccc@{}}
		\toprule
		                            & \multicolumn{2}{c}{\textbf{POS}} & \multicolumn{2}{c}{\textbf{DEP}}                                   \\ \cmidrule(lr){2-3} \cmidrule(lr){4-5}
		\textbf{Encoder / treebank} & \textbf{Layer}                   & \textbf{Aggr.}                   & \textbf{Layer} & \textbf{Aggr.} \\
		RoBERTa / en\_gum           & 3                                & max                              & 3              & mean           \\
		XLM-R / en\_gum             & 9                                & max                              & 9              & first          \\
		XLM-R / it\_vit             & 9                                & first                            & 9              & mean           \\
		XLM-R / el\_gdt             & 12                               & mean                             & 9              & mean           \\ \bottomrule
	\end{tabular}
	\caption{Optimal combination of embedding layer and subword pooling function for each encoder
		(RoBERTa \& XLM-R), task (POS \& DEP) and language (English, Italian \& Greek) combination,
		chosen as outlined in Section~\ref{sec:probe_config} and used throughout our experiments. When
		neutralizing across languages (Section \ref{sec:cross-lang}) and across tasks (Section
		\ref{sec:cross-task}), we use the configuration of the neutralizer for both neutralizer and
		target.}
	\label{tab:configurations}
\end{table}

\subsubsection{Models}

\paragraph{Encoders} We use RoBERTa~\citep{liu_roberta_2019} and
XLM-R~\citep{conneau_unsupervised_2020} as the encoders we probe. RoBERTa is an optimized version of
the encoder-only transformer BERT~\citep{devlin_bert_2019}, and XLM-R is its multilingual variant,
supporting 100 languages. Unlike BERT, both models are trained exclusively on the masked language
modeling (MLM) objective. We use the ``base'' version of each model, comprising of 12 encoder
layers, 12 attention heads and an embedding size of 768, with a total parameter count of $\sim$125
million\footnote{\cameraready{We report the results of a brief parameter scaling experiment in
		Appendix \ref{app:scale}.}}.

\paragraph{Probing classifier} \cameraready{For our probing classifier, we use a multilayer
	perceptron (MLP) consisting of two linear layers with a $\tanh$ activation function in between}. We
train our probing classifier using the AdamW optimizer~\citep{loshchilov_decoupled_2019} with
a learning rate of $10^{-3}$ and weight decay of $10^{-2}$, and employ an early stopping criterion
based on the validation set. We report the classification accuracy for the best configurations in
Table~\ref{tab:baseline_acc}.

\cameraready{\subsection{Probing Classifier Selectivity Baseline}

  A sufficiently expressive probing classifier may be capable of learning \emph{any} task given
  representations as input. This would render its probing functionality obsolete since the
  information localized by the probing classifier would no longer be necessarily attributed solely
  to the input representations. To ensure that our probing classifier is not overly expressive, we
  conduct a baseline check as outlined by \citet{hewitt_designing_2019}. Here, we construct an
  analogous \textit{control task} for POS tagging, where we randomly assign each word in the
  training set to one of $N$ arbitrary labels, where $N$ is the number of POS tag labels (i.e. 17).
  A good probe should have high \textit{selectivity}, computed as the difference between the
  accuracy on the probing task and the accuracy on the control task. The more selective the probe,
  more likely it is that the information it accesses is specific to the input representations. If
  the probe performs equally well on the control task, it suggests that the probe may be leveraging
  some inherent properties of the model architecture or the data distribution, rather than
  specifically extracting useful information from the representations. We train a new checkpoint of
  our probing classifier on the control task, ending training after the same number of update steps
performed when training on the POS tagging task, and report the outcome in Section
\ref{sec:joint-pos}.}

\subsection{Centroid estimation}\label{sec:centroid_estimation}

To study the joint encoding of linguistic categories (e.g. nouns, verbs, etc.) in our encoders, we
use the probing classifier to localize the subspace of the encoder that corresponds to each of these
categories by obtaining their mean vector representation, or ``centroid'', similarly to the language
centroids of \citet{libovicky_language_2020}. The intuition is that representations that repeatedly
result in predictions of a particular class will encode information specific to that class, which
can be captured through aggregation such as averaging. For POS tagging, the centroid $u_t$ of each
target POS class $t$ is defined as
\begin{equation}
	\mathbf{u}_{t}^{(POS)} := \frac{1}{|V_t^l|} \sum_{\mathbf{v} \in V_t^l} \mathbf{v}
\end{equation}
where $V_t^l$ is the set of the representations of the words in our validation set that were
predicted as tag $t$ when probing the $l$th layer of our the encoder. For DEP labeling, each
prediction depends on both the representation of the head $\mathbf{h}$ and the child $\mathbf{c}$ of
the DEP relation. The centroid of each target DEP relation $t$ is computed as:
\begin{equation}\label{eq_dep}
	\mathbf{u}_{t}^{(DEP)} := \frac{1}{|P_t^l|} \sum_{(\mathbf{h}, \mathbf{d}) \in P_t^l} [\mathbf{h} ~; \mathbf{d}]
\end{equation}
where $P_t^l$ is the set of (head, dependent) representation pairs that were predicted as DEP
relation $t$ when probing the $l$-th layer of our encoder, and $[\mathbf{h} ~; \mathbf{d}]$ is the
concatenation of the representations.\footnote{We test the quality of the predicted centroids by
	computing gold centroids based on the gold labels, and computing cosine similarity between the
	two. We find that they are near identical ($\sim$1 similarity), see Appendix
	\ref{app:golden_centroids}.}

\subsection{Cross-neutralization}

To study whether different linguistic categories are jointly encoded within LLMs, we use
a cross-neutralization method. We first evaluate the class-specific accuracy of our probing
classifier on the original representations of our encoder. We then take a \texttt{neutralizer}
centroid estimated as outlined in Section \ref{sec:centroid_estimation}, and subtract it from the
encoder representations corresponding to some \texttt{target} category. For instance, taking
\emph{verbs} as neutralizer and \emph{nouns} as target, we subtract all \emph{verb} information from
the \emph{noun} representations. We then repeat probing classification on these ``neutralized''
encoder representations. The intuition behind this is that if the encoders were to represent
linguistic categories in independent ways, we expect the performance to deteriorate only for the
linguistic category that we use for computing the centroids, i.e. removing \emph{verb} information
only negatively affects the representations of \emph{verbs}. However, in the case of joint encoding,
we expect to see substantial performance drops for other \texttt{target} categories as well,
suggesting that some features of the encoder play a role in encoding both categories.
\cameraready{As a baseline, we additionally experiment with subtracting random vectors (rather than
	centroids) from the encoder representations. If the centroids are indeed responsible for the drop in
	performance, then we should not observe similar performance drops when subtracting random vectors.
	We report the outcome of this experiment in Section \ref{sec:joint-pos}.}

\subsection{Choosing a configuration for probing}\label{sec:probe_config}

Understanding where to best localize the task-specific information within the models is not trivial.
For instance, while \citet{tenney_bert_2019} show that syntax information is more localized in lower
layers, \citet{de2020s} demonstrate that such findings do not automatically port to multilingual
models. Moreover, \citet{del2022similarity} show that for multilingual models, the pooling method
used can have important effects on the knowledge that is captured. Thus, we first study which layer
$l$ and pooling function maximizes the amount of information that is captured in our centroids. As
such, we employ the notion of \textit{self-neutralization}, the case in which the same linguistic
category is both the target and the neutralizer.

We hypothesize that a large drop in accuracy after self-neutralization is indicative of a high
amount of relevant information being captured by the centroid. This drop should be relative to
a high accuracy. We thus select the top quartile of our configurations in terms of original-encoder
accuracy, and from this subset use the configuration with the highest relative drop due to
self-neutralization. We consider representations from layers $l \in \{1, 3, 6, 9, 12\}$ and pool
subwords by either taking the first token in a word, or by max-pooling or mean-pooling over the
token representations. For each encoder, language and task, we separately find the optimal probing
configuration and report them in Table~\ref{tab:configurations}. We make use of the validation set
for this portion of the methodology.

\begin{figure}[!t]
	\centering
	\includegraphics[width=0.9\columnwidth]{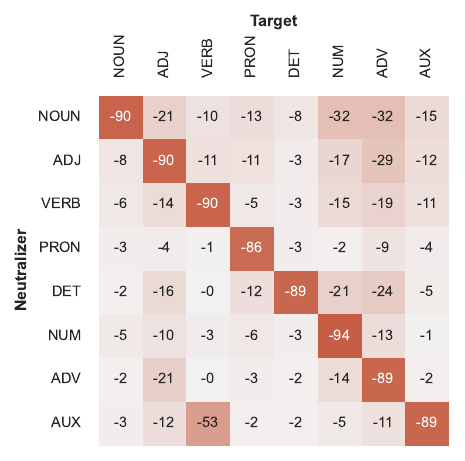}
	\caption{Relative change in accuracy when cross-neutralizing POS tags (RoBERTa).}
	\label{fig:xn-roberta}
\end{figure}

\section{Joint encoding of POS tags}\label{sec:joint-pos}

\begin{figure*}[!t]
	\centering
	\includegraphics[width=0.95\textwidth]{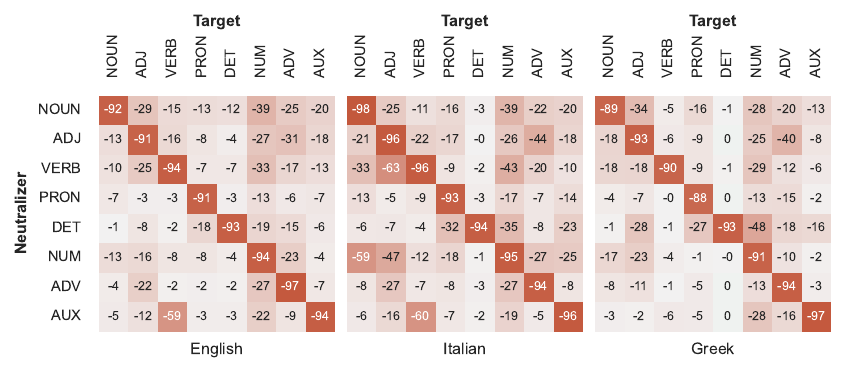}
	\caption{Relative change in accuracy when cross-neutralizing POS tags using embeddings from XLM-R in
		(left-to-right respectively) English, Italian and Greek, labeled by their universal dependencies
		treebank.}
	\label{fig:xn-xlm_r}
\end{figure*}

We begin by examining whether representations of different POS tags share information. For each POS
tag, we compute its centroid and re-classify representations neutralized by the centroid.
\cameraready{With regard to our baselines, we find that subtracting random vectors leaves the
	performance relatively unchanged, with no evidence of systematic drops. For our selectivity
	baseline, we find that our probing classifier achieves an average accuracy of 96\% on the POS
	tagging task and 63\% on the control task, giving a selectivity value of 33\%. This is in line
	with probes for similar tasks accepted for their validity in the
	literature~\citep{hewitt_designing_2019}. }

To facilitate visualization and discussion, we select an illustrative subset\footnote{See
	Appendix~\ref{app:detailed-xn} for results between all classes.} of POS tags that covers both open-
and closed-class words, as we expect potentially different patterns between these two groups. For
instance, we expect neutralization to be symmetric (neutralizer and target can be swapped for
similar results) in the case of open-class words as we hypothesize that joint encoding here is
dictated by co-occurrence patterns. Conversely, we expect asymmetric neutralization in the case of
close-class neutralizers, supposing that joint encoding here may be dictated by functional
dependencies, which may not necessarily be reciprocal.

\paragraph{Results}

We find that linguistically related categories generally tend to be jointly encoded in RoBERTa (see
Fig.~\ref{fig:xn-roberta}). For instance, the relative decrease of \texttt{VERB} classification
accuracy by $53\%$ when neutralizing using auxiliaries (\texttt{AUX}) suggests information sharing.
This may be explained by the fact that auxiliaries themselves are verbs (e.g. ``\textit{has}
done''), and that they functionally modify verbs. We note that the information flow in this pairing
is asymmetric, as noted by the much smaller $11\%$ drop in the reciprocal case. This aligns with our
hypothesis that joint encoding for closed-class neutralizers may be explained by their functional
role, which is often not reciprocal (verbs will generally not modify auxiliaries).

While we observe further evidence of linguistically related categories being jointly encoded (e.g.
\texttt{VERB} and \texttt{ADV}), this is not ubiquitous. For instance, joint encoding is not found
when neutralizing nouns (\texttt{NOUN}) with determiners (\texttt{DET}). Here, the relative
percentage decrease is only $2\%$, despite determiners acting as modifiers for nouns. In general, we
do not find distinct patterns of joint encoding unique to closed-class and open-class words. Our
hypothesis of symmetry due to co-occurrence is at best supported by the \texttt{ADV}-\texttt{ADJ}
pairing, with relative drops of -21\% and -29\%. In most other pairings information sharing appears
to be asymmetric, suggesting functional dependence or another root cause as explanation.

Moreover, we note that open-class words tend to be ``adept'' at cross-neutralizing, sharing
information as neutralizers with many other tags as shown by the \texttt{NOUN}, \texttt{ADJ} and
\texttt{VERB} neutralizers rows successfully neutralizing several columns. However, we note similar
patterns when neutralizing with the closed-class \texttt{DET}. Overall, we find evidence of joint
encoding of POS tags in RoBERTa, but further work is necessary to establish which mechanisms lead to
specific pairs sharing information or not.

\subsection{Joint encoding in multilingual models}\label{sec:multilingual}

We now investigate how our findings transfer to other languages. We examine to what extent
multilingual models demonstrate language-specific or language-agnostic information-sharing behavior.
Additionally, we verify the consistency of our method across models as we can directly compare the
results between RoBERTa and XLM-R in English, and also across languages, as we can compare the
effects of cross-neutralization on English, Italian, and Greek. Our setup is the same as in Section
\ref{sec:joint-pos}, replacing RoBERTa with XLM-R and repeating the experiment in the three
languages.

\paragraph{Monolingual vs. multilingual model}\label{sec:mono-vs-multi}
When comparing RoBERTa and XLM-R on the English dataset, we observe similar patterns across POS tags
(see Fig. \ref{fig:xn-roberta} and the left-most column of Fig. \ref{fig:xn-xlm_r}). For instance,
for both models, we see that nouns have a neutralization effect on adjectives, numerals, and
adverbs. This suggests that given a language,  multilingual models jointly encode POS tags similarly
to their monolingual counterparts.

\paragraph{Joint encoding within languages}\label{sec:multilingual-res}
We find that XLM-R exhibits similar sharing patterns within different languages (see Fig.
\ref{fig:xn-xlm_r}). For instance, we see that adjectives consistently neutralize adverbs across all
languages (-31\%, -44\% and -40\% in English, Italian and Greek). This suggests that the
representations responsible for predicting these tags contain some language-agnostic information.

However, we also observe some language-specific behavior. For instance, \texttt{VERB} categories
cross-neutralize adjectives (\texttt{ADJ}) more prominently in Italian (-63\%) than in English
(-26\%) and Greek (-18\%). This may be explained by the postnominal use of adjectives in Italian
which may overlap more closely with the positioning of verbs, particularly in the past participle
tense~\citep{cinque_syntax_2010} (in contrast to the strictly prenominal use in the other
languages). The results suggest that XLM-R discovers and leverages language-agnostic information
when possible, while also learning language-specific information when necessary.

\section{Information sharing across languages}\label{sec:cross-lang}

Our findings for individual languages on XLM-R raise the question of whether information about
linguistic categories from two \textit{different} languages can also be jointly encoded. To further
probe for language-agnostic joint encoding of linguistic categories, we extend our experiment to
test whether information between two linguistic categories from two different languages is jointly
encoded. We posit that if categories share information across languages, this is evidence that the
models learn at least partially language-independent representations for these categories. We test
this hypothesis by cross-neutralizing every linguistic category in a language $A$ with another
category from a language $B$, e.g. neutralizing all Italian POS tags with English nouns. Aside from
probing for language-agnosticism, this allows for further exploration of joint encoding of
linguistic categories. To enable consistent neutralization, we use the same layer and pooling
function configuration (see Section \ref{sec:probe_config}) for both neutralizer and target. We do
this by utilizing the neutralizing language configuration for both neutralizer and target in cases
where they differ.

\begin{figure}[!t]
	\centering
	\includegraphics[width=\columnwidth]{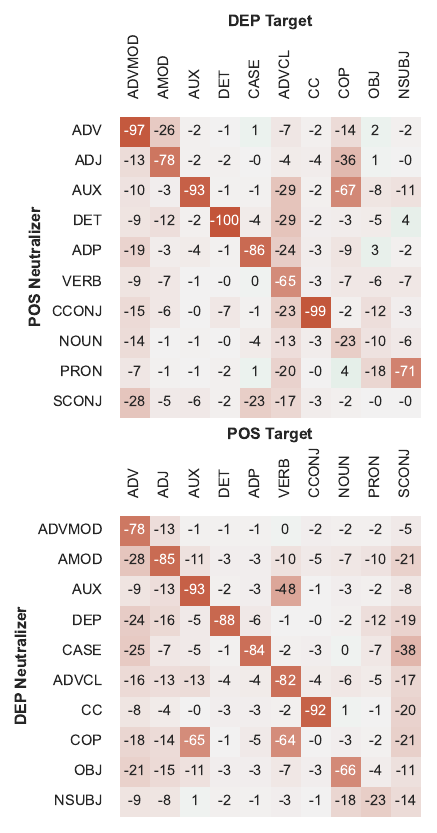}
	\caption{Relative change in accuracy cross-neutralizing RoBERTa DEP representations using RoBERTa
		POS centroids (top) and vice versa (bottom)\protect\footnotemark.}
	\label{fig:dep-xt-xn-posneutr-roberta}
\end{figure}
\footnotetext{Note that because the linguistic categories are now different between neutralizer and
	target, the diagonal no longer represents self-neutralization, and is simply an artifact of the
	ordering of the categories to favor legibility.}

\begin{figure}[!t]
	\centering
	\includegraphics[width=0.9\columnwidth]{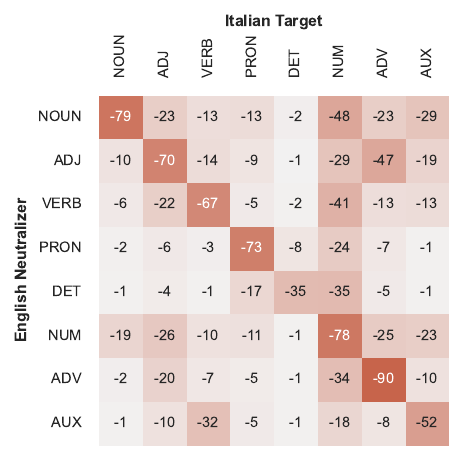}
	\caption{Relative change in accuracy when cross-neutralizing XLM-R embeddings in Italian using
		centroids from English.}
	\label{fig:xl-xn}
\end{figure}

\paragraph{Results}\label{sec:cross-lang-res}
Fig.~\ref{fig:xl-xn} shows results of cross-neutralizing for Italian POS tags when using the
corresponding English centroids from each linguistic category. On the diagonal, neutralizer and
targets correspond to the same linguistic category, but for different languages. Accuracy drops
substantially here ($\sim70\%$ on avg.). This is evidence of language-agnostic encoding of these
linguistic categories, as centroids in one language neutralizing targets in another language points
to information in the representations being shared across both languages.

\begin{figure*}[!t]
	\centering
	\includegraphics[width=\textwidth]{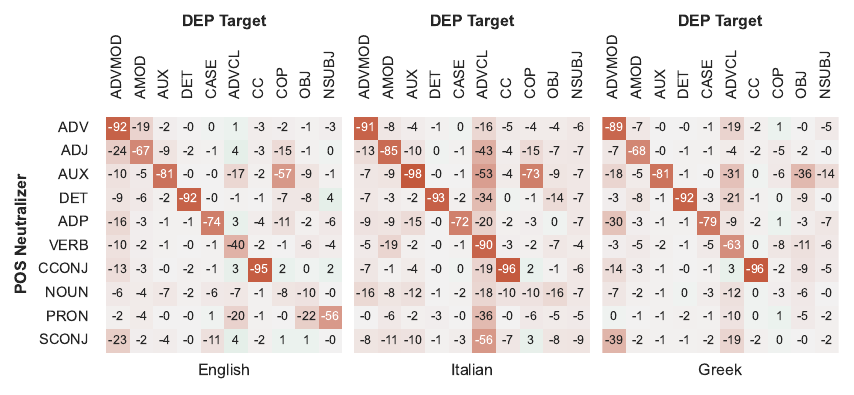}
	\caption{Relative change in accuracy for a sample of dependency relations when cross-neutralizing
		XLM-R DEP embeddings using XLM-R POS centroids in (left-to-right respectively) English, Italian
		and Greek.}
	\label{fig:dep-xt-xn-posneutr-xlm_r}
\end{figure*}

We also observe joint encoding of different linguistic categories, for instance with English
\texttt{NOUN}s neutralizing Italian numerals (\texttt{NUM}) (-48\%), or English adjectives
(\texttt{ADJ}) neutralizing Italian adverbs (\texttt{ADV}) (-47\%). An alternative explanation to
language-agnostic representation learning in XLM-R may be that instances of code-switching or
language borrowing in the pretraining data may have encouraged XLM-R to jointly encode different
linguistic categories across languages. Future work may investigate why joint encoding occurs across
different categories of different languages.

We repeat the experiment using English neutralizers on the Greek corpus, and observe similar but
milder trends. This may be due to English being phylogenetically closer to Italian than
Greek~\citep{chang_ancestry-constrained_2015}, leading to a higher degree of information sharing
between the former two languages, but further work in determining which language pairs share more
information is needed to draw solid conclusions. For more detailed plots on cross-lingual
cross-neutralization, we refer the reader to Appendix~\ref{app:detailed-xn}, where we present
results from all combinations of neutralizer and target across the three languages.

\section{Information sharing across tasks}\label{sec:cross-task}

To study whether information is shared across the linguistic hierarchy, we cross-neutralize between
tasks using POS neutralizers and dependency relation targets. Since a given parent may have several
dependents while the child will only have one parent, we posit that most of the information for the
child-parent dependency is captured by the child. We therefore subtract the POS centroids from the
child representations in the child-parent concatenations for dependency labeling. We also complete
the complementary experiment of subtracting the child portion of the dependency relation centroids
from the POS embeddings for POS tagging, to test whether joint encoding happens in both directions.
As in Section \ref{sec:cross-lang}, we use the neutralizer configuration for both neutralizer and
target for consistency.

\paragraph{Results}\label{sec:cross-task-res} Fig. \ref{fig:dep-xt-xn-posneutr-roberta} (top) shows
the results when neutralizing RoBERTa representations for DEP labels using POS centroids. We find
further evidence of linguistically related units being jointly encoded. For instance, adverb POS
tags (\texttt{ADV}) neutralize adverbial modifier (\texttt{ADVMOD}) dependency labels, as can be
seen by the 97\% relative drop in accuracy. The fact that POS tag representations jointly contain
information that is crucial for encoding DEP labels shows that LLMs learn hierarchically,
reminiscent of the classical NLP pipeline.

To study whether our findings generalize beyond English and RoBERTa, we repeat the experiment on
XLM-R (see Fig. \ref{fig:dep-xt-xn-posneutr-xlm_r}). Large drops in accuracy on the diagonal suggest
that many information-sharing patterns hold across languages and models, consolidating our findings
from sections~\ref{sec:multilingual} and~\ref{sec:cross-lang}. We also find language-specific
results: pronouns (\texttt{PRON}) neutralize nominal subjects (\texttt{NSUBJ}) only in English
(-56\%), having surprisingly little effect on the other languages (-5\% and -2\%). On occasion, we
find evidence of bidirectional sharing between different levels of the linguistic hierarchy. In Fig.
\ref{fig:dep-xt-xn-posneutr-roberta} (bottom), we note that certain pairs highlighted by the
diagonals such as adpositions (POS: \texttt{ADP}) and case relations (DEP: \texttt{CASE}) present
evidence of joint encoding both when neutralizing DEP with POS centroids (-55\%) and when
neutralizing POS with DEP centroids (-84\%). This seems to suggest that the hierarchical nature of
the representations learned by these LLMs is not necessarily unidirectional: information appears to
be shared both upwards and downwards in the linguistic hierarchy.

\section{Conclusion}

We study information sharing between linguistic categories in LLMs, finding evidence of joint
encoding between pairs of related POS tag classes. By applying our method to XLM-R, we find evidence
of joint encoding in XLM-R across languages, showing that our findings hold for both the monolingual
(RoBERTa) and multilingual (XLM-R) case. Lastly, we cross-neutralize between POS tagging and DEP
labeling, and find evidence of information sharing across the linguistic hierarchy. \cameraready{We
	test specifically for joint encoding of different syntactic categories that rely on the same
	linguistic concept, such as an “amod” dependency implying some relationship with the adjective and
	the noun POS tags. However, our method could be extended to test for joint encoding of categories in
	other tasks that can be expected to share information. This in turn could be informative on whether
	an LLM indeed captures this shared information between different tasks. A more complete map of this
	knowledge could help develop better models in many application scenarios: for instance, in
	a multitask learning setting where negative interference between parameter updates from different
	tasks is known to hamper performance~\citep{zhao_modulation_2018} or a lifelong learning setting
	where learning a new task often leads to (catastrophic) forgetting of the previously learned
	ones~\citep{biesialska_continual_2020}.} Future work may additionally seek to explore more
languages, particularly low-resource and typologically distant languages, or consider including
higher-level semantic tasks from the classical NLP pipeline, such as semantic role labeling, word
sense disambiguation and coreference resolution. Further work is also necessary for understanding
\textit{why} certain representations share information while others do not.

\section*{Limitations}

In this work, we limit our experiments to an encoder-only architecture. Further research could be
carried out on encoder-decoder or decoder-only architectures.

Additionally, our experiments only examine three languages. Future work may therefore aim to
extend the experiments to include more languages, particularly low-resource languages.
Similarly, our discussion of the linguistic hierarchy is limited to two tasks from the early
stages of the classical NLP pipeline. As mentioned, further work may extend the experiments to
higher-level semantic tasks.

Lastly, the lack of cross-neutralization effects between some linguistic categories does not
necessarily indicate the absence of joint encoding, but merely that our representations were not
sufficient to prove its existence.

\bibliography{main}
\bibliographystyle{acl_natbib}

\clearpage
\appendix

\section{Dataset pre-processing}
\label{app:dataset_preprocess}

Table \ref{tab:datasets} presents the train, validation and test split sizes (in terms of number of
sentences) for each language considered in our work.

The sentences in the corpora from the Universal Dependencies framework are already tokenized to the
word level and stored as lists of words in the \texttt{tokens} field. However, since we use sub-word
tokenizers, namely the Byte Pair Encoding~\citep{sennrich_neural_2016} and
SentencePiece~\citep{kudo_sentencepiece_2018} tokenizer, we further split the words into their
sub-word tokens. Depending on the task, we also include either the \texttt{upos} field, which is
a list of integers corresponding to one of the 17 universal POS tags available, or the \texttt{head}
and \texttt{deprel} fields which contain the head and one of the 36 dependence relations for
dependency labeling\footnotemark. It should be noted that we only keep the language-independent
relations, as some of them appear only with a language-specific modifier, and including them would
make comparisons across languages less straightforward\footnote{A full list of
	\href{https://universaldependencies.org/u/pos/all.html}{POS tags} and
	\href{https://universaldependencies.org/u/dep/all.html}{dependency relations} can be found on the
	Universal Dependencies website.}.

Furthermore, upon inspecting the datasets we observed that the annotators had split contractions
into their parts and included them next to the original contraction for the Italian and Greek
corpora. However, ground-truth labels were only provided for the sub-words, with the compound words
annotated as a special class ``\_''. Hence, we filtered out the compound words from these datasets
and retained their sub-parts. In addition to that, for dependency labeling, we ignored words with
the root dependency label, as they have no head and their prediction is trivial.

\cameraready{\section{Scaling to larger models}\label{app:scale} We repeat the experiments from
	Sections \ref{sec:joint-pos} \& \ref{sec:cross-lang} to verify whether they hold across different
	model sizes. More specifically, we scale the English POS tag cross-neutralization from
	RoBERTA-base to RoBERTA-large (Fig.~\ref{fig:xn-roberta-large}), and from XLM-R-base to
	XLM-R-large (Fig.~\ref{fig:xn-xlmr-large}) and XLM-R-XL (Fig.~\ref{fig:xn-xlmr-xl}). We notice
	that similar patterns occur across all tested model sizes (e.g. \texttt{NOUN} neutralizing
	\texttt{ADV}). The effect appears to be generally less pronounced for larger models, although we
	also did find instances where the drop in accuracy is the same or higher. }

\section{Golden and Predicted Centroid Similarity}\label{app:golden_centroids}

We present the cosine similarity between golden and predicted centroids for the POS tagging setup in
table \ref{tab:golden_centroids}, as mentioned in Section \ref{sec:centroid_estimation}.

\section{Self-neutralization visualization}
\label{app:self-neutr}

\begin{figure*}[thb]
	\centering
	\begin{subfigure}{\columnwidth}
		\centering
		\includegraphics[width=.9\textwidth]{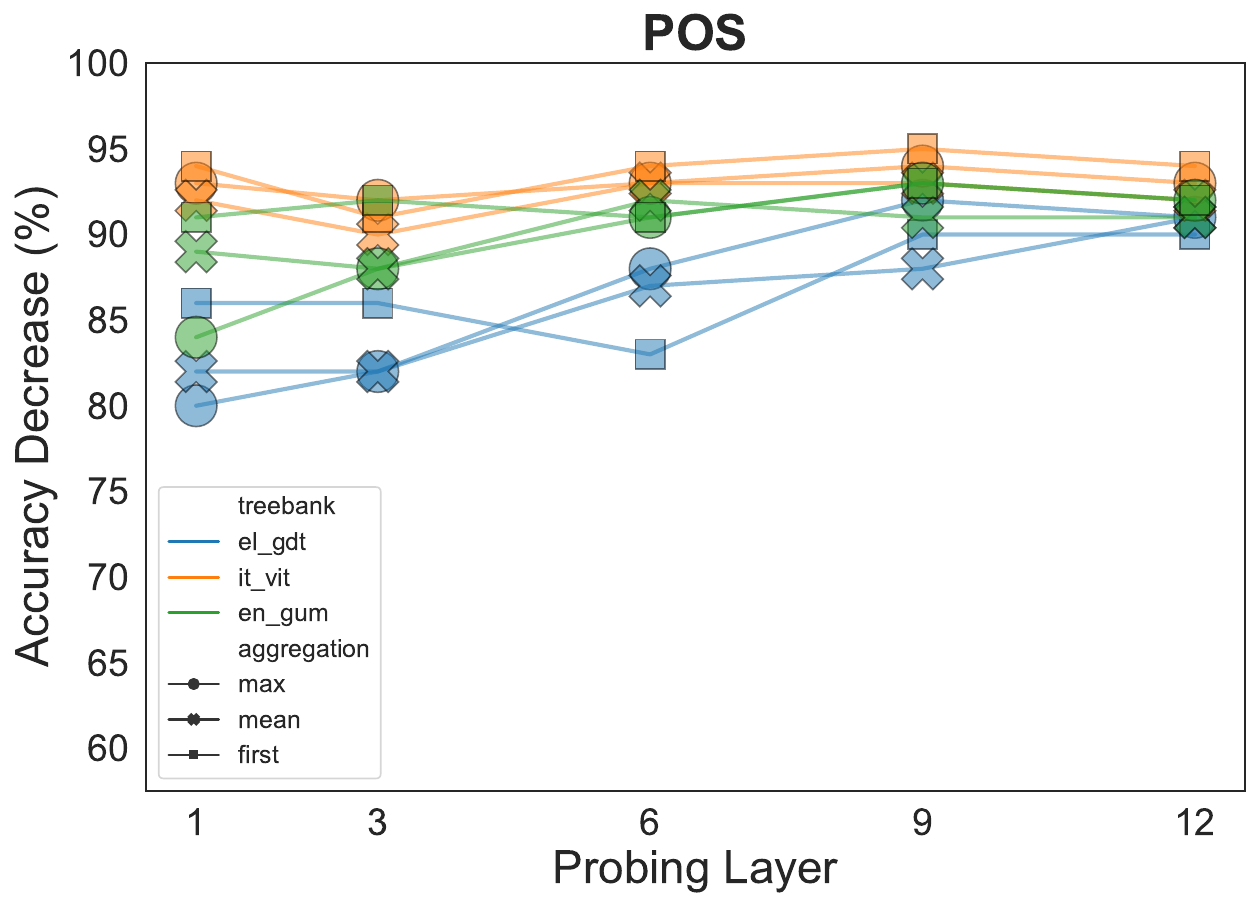}
		\caption{POS tagging Accuracy Decrease using different WordPiece pooling methods
			(``aggregation'' in the plot).}
		\label{fig:pos_selection_xlm}
	\end{subfigure}
	\begin{subfigure}{\columnwidth}
		\centering
		\includegraphics[width=.9\textwidth]{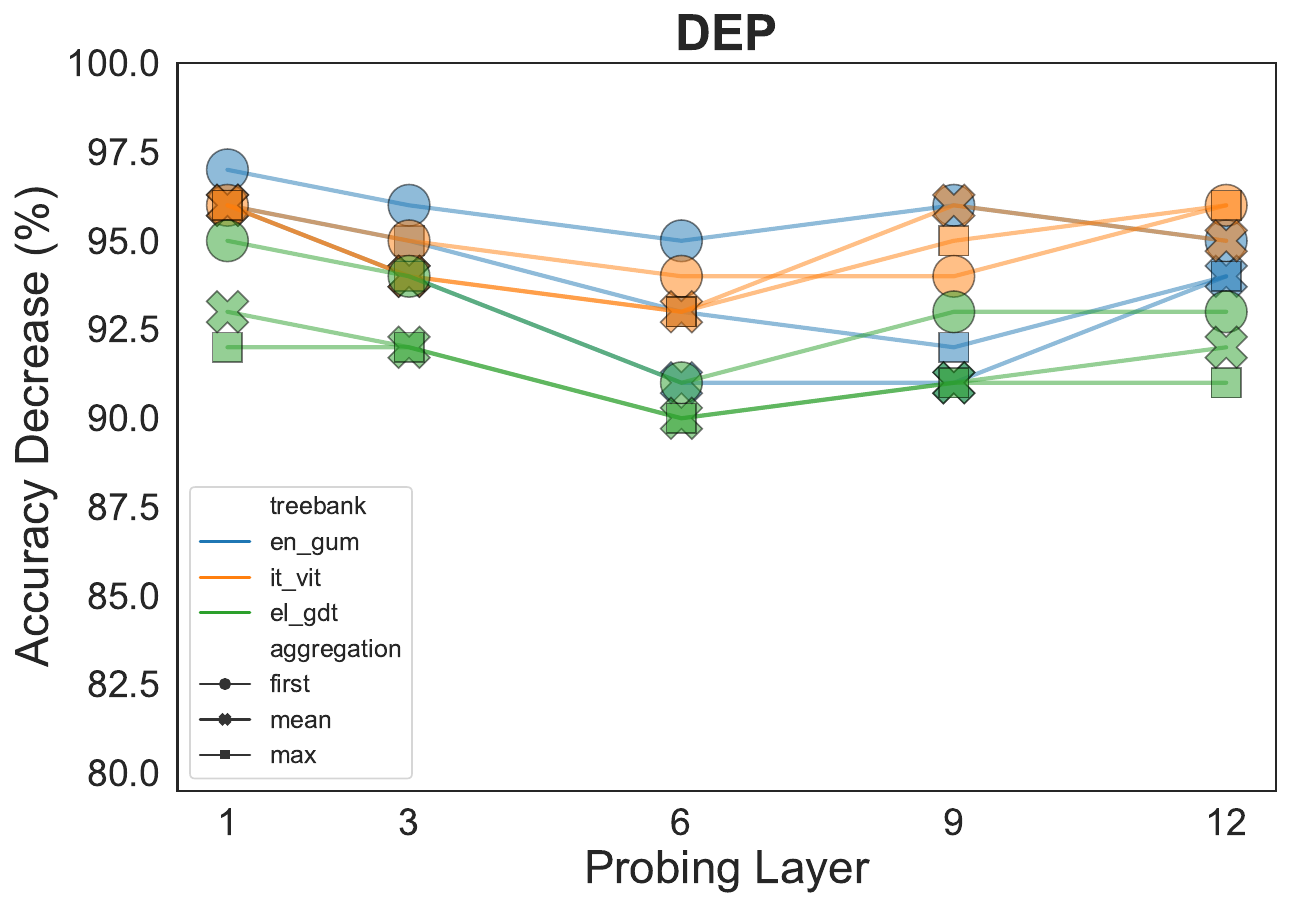}
		\caption{Dependency labeling Accuracy Decrease using different WordPiece pooling methods
			(``aggregation'' in the plot).}
		\label{fig:dep_selection_xlm}
	\end{subfigure}
	\caption{Decrease in performance for XLM-R when \textbf{self-neutralizing} in the POS tagging
		(left) and dependency labeling (right) tasks using embeddings extracted from different layers
		and setup configurations, for each of for English (en\_gum), Italian (it\_vit) and Greek
		(el\_gdt) treebanks. For dependency labeling, we use the best child-head concatenation mode
		(ONLY) based on the results we aquired with RoBERTa, as shown in
		Fig.~\ref{fig:dep_selection_roberta}.}
	\label{fig:selection_xlm}
\end{figure*}

\begin{figure}[htb]
	\centering
	\begin{subfigure}{\columnwidth}
		\centering
		\includegraphics[width=\textwidth]{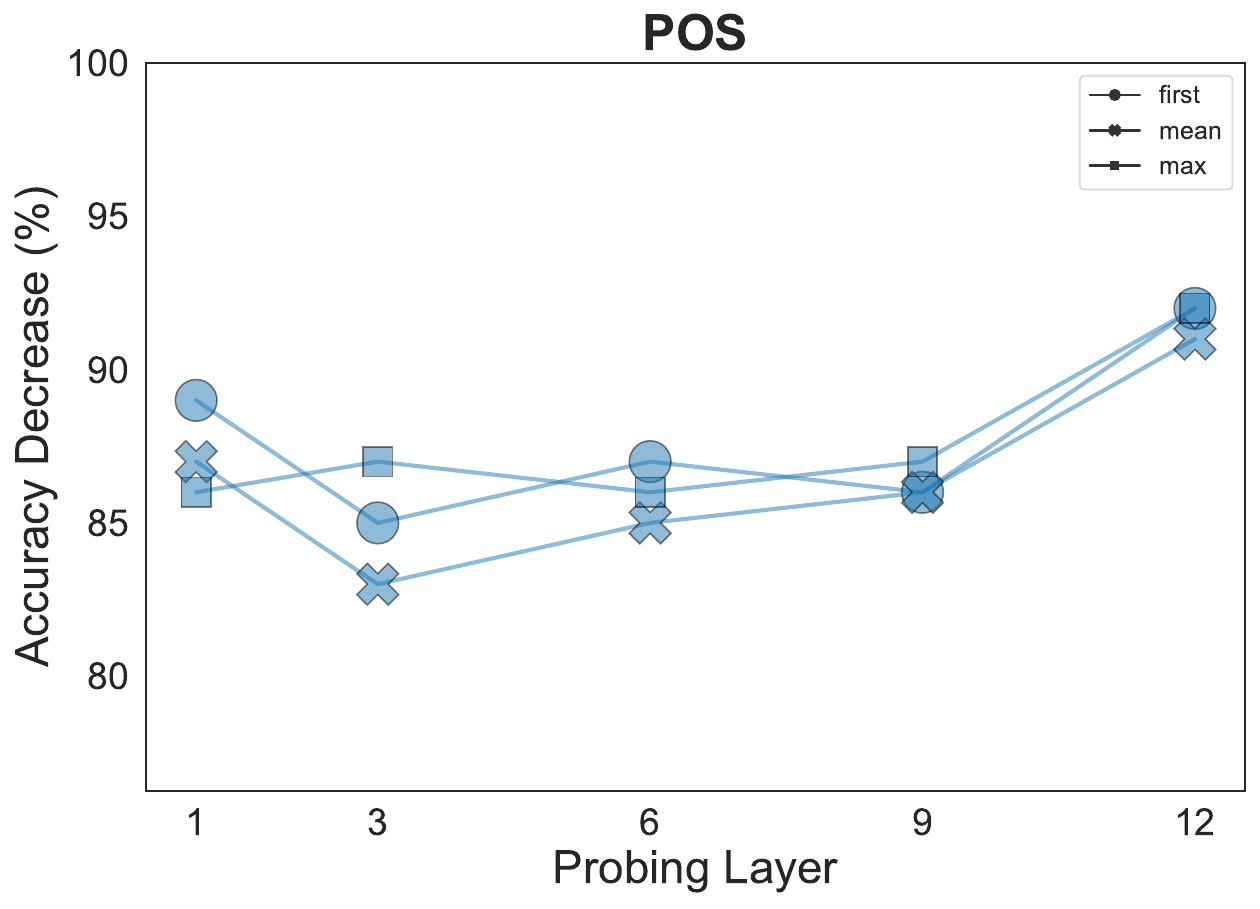}
		\caption{Accuracy decrease for \textbf{POS tagging} when using different WordPiece
			poolings.}
		\label{fig:pos_selection_roberta}
	\end{subfigure}
	\begin{subfigure}{\columnwidth}
		\centering
		\includegraphics[width=\textwidth]{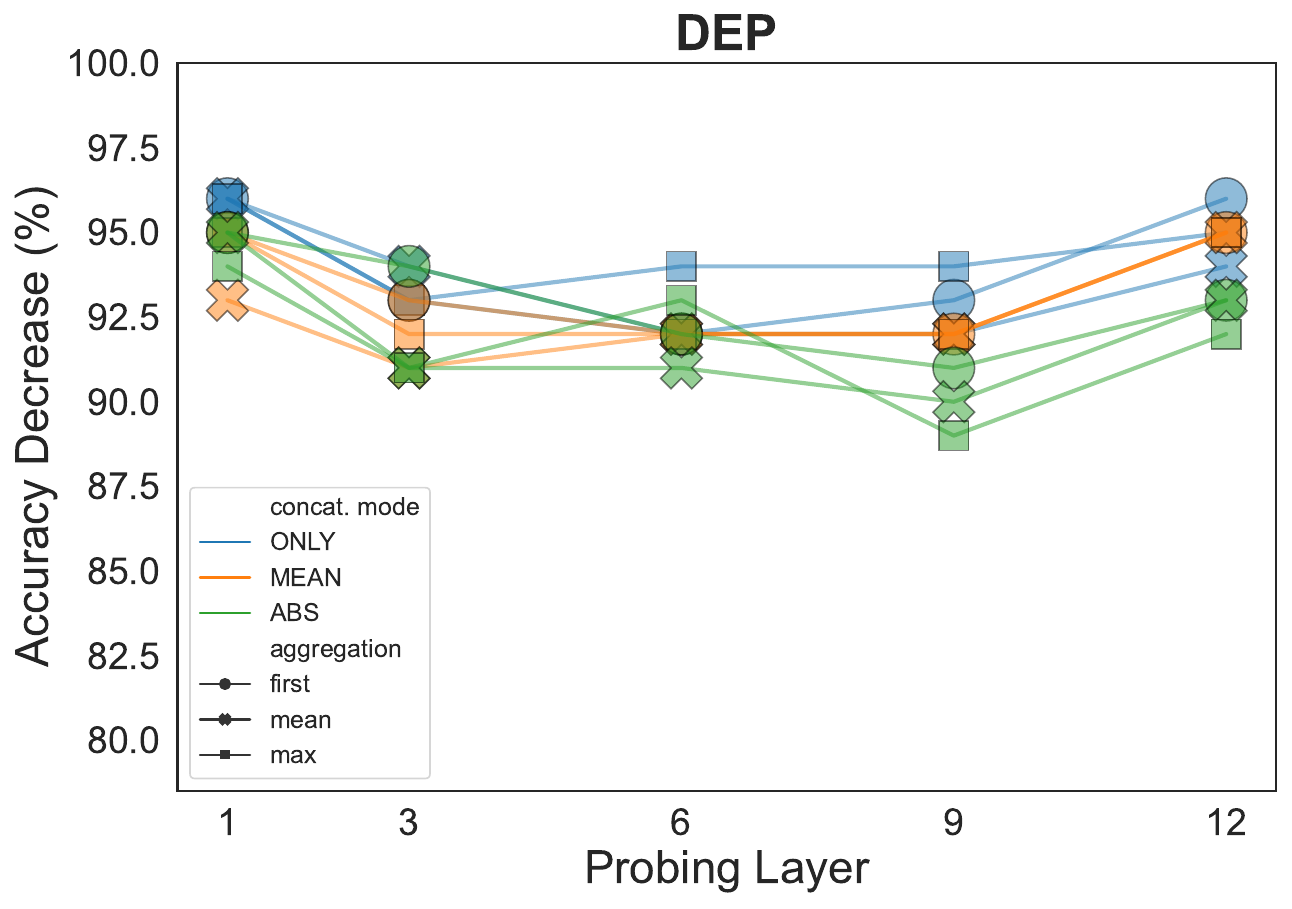}
		\caption{Accuracy decrease for \textbf{dependency labeling} when using different WordPiece
			pooling methods (``aggregation'' in the plot) and child-head concatenation configurations.}
		\label{fig:dep_selection_roberta}
	\end{subfigure}
	\caption{Decrease in performance for RoBERTa when \textbf{self-neutralizing} in the POS tagging
		(top) and dependency labeling (bottom) tasks using embeddings extracted from different layers
		and setup configurations.}
	\label{fig:selection_roberta}
\end{figure}

Figs.~\ref{fig:selection_roberta} and~\ref{fig:selection_xlm} showcase the decrease in accuracy when
self-neutralizing in the POS tagging/dependency labeling task for RoBERTa and XLM-R accordingly.

\section{Engineering Logistics}

We rely on PyTorch Lightning~\citep{falcon_pytorch_2019} and Hugging Face Datasets
\citep{lhoest_datasets_2021} and Transfomers~\citep{wolf_huggingfaces_2020} for our implementation.
We run our experiments on an NVIDIA A100 GPU, with each training run taking approximately three
minutes. Our code is available at
\href{https://github.com/thesofakillers/infoshare}{github.com/thesofakillers/infoshare}.

\section{Detailed Cross-Neutralizing Results}
\label{app:detailed-xn}

\begin{figure*}[ht]
	\centering
	\includegraphics[width=\textwidth]{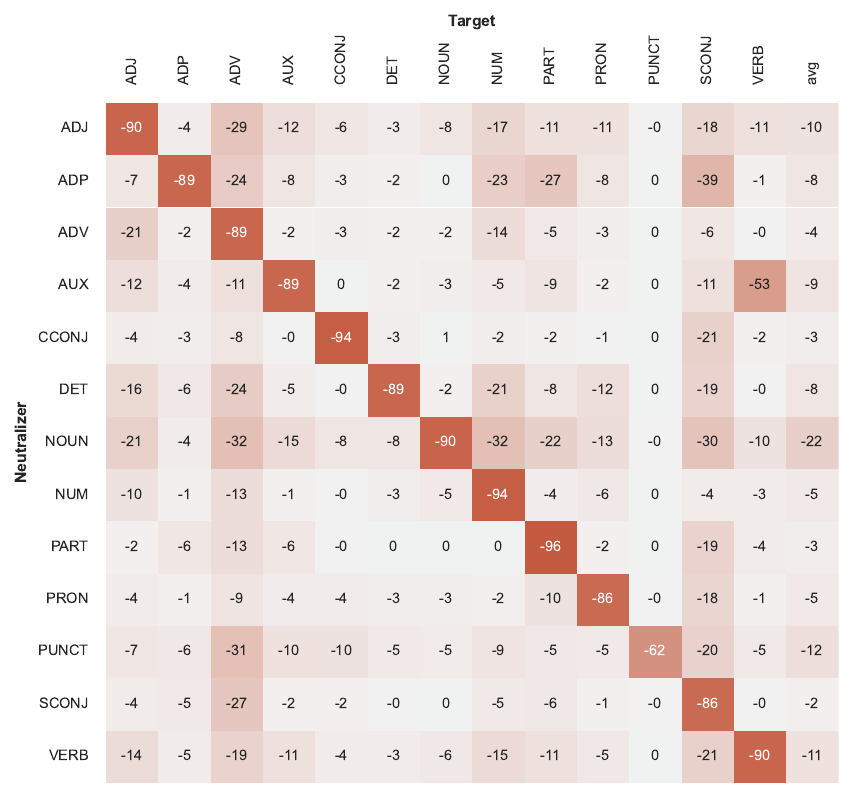}
	\caption{Relative change in accuracy when cross-neutralizing POS tags using centroids from RoBERTa
		in English.}
	\label{fig:xneutr_roberta_pos_complete}
\end{figure*}

\begin{figure*}[t]
	\centering
	\begin{subfigure}{.32\textwidth}
		\centering
		\includegraphics[width=\textwidth]{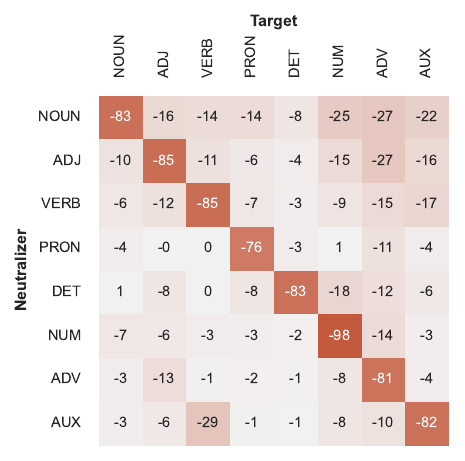}
		\caption{RoBERTa-large}
		\label{fig:xn-roberta-large}
	\end{subfigure}
	\begin{subfigure}{.32\textwidth}
		\centering
		\includegraphics[width=\textwidth]{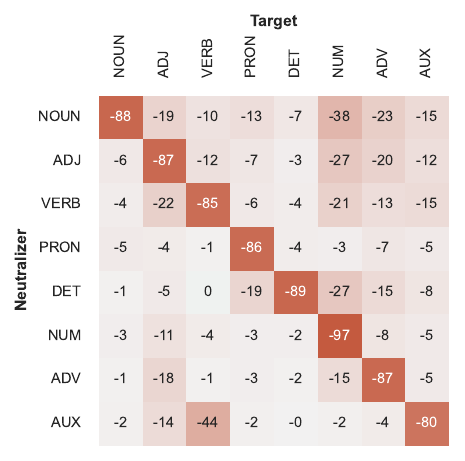}
		\caption{XLM-R-large}
		\label{fig:xn-xlmr-large}
	\end{subfigure}
	\begin{subfigure}{.32\textwidth}
		\centering
		\includegraphics[width=\textwidth]{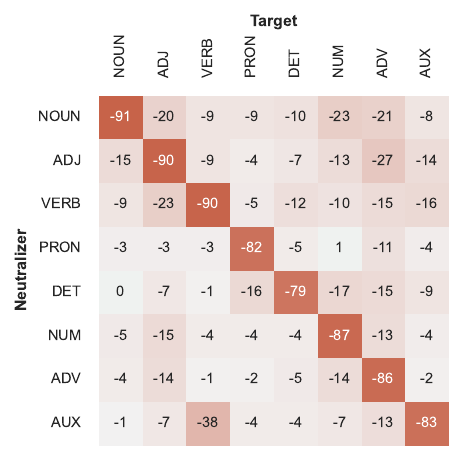}
		\caption{XLM-R-XL}
		\label{fig:xn-xlmr-xl}
	\end{subfigure}
	\caption{Change in accuracy when cross-neutralizing English POS tags with centroids from different
		model sizes.}
\end{figure*}

\begin{figure*}[ht]
	\centering
	\includegraphics[width=0.49\textwidth]{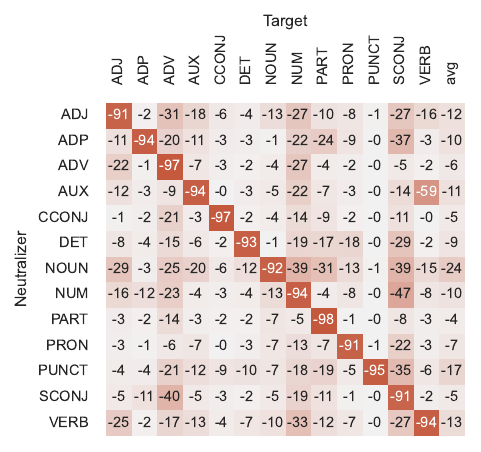}
	\includegraphics[width=0.49\textwidth]{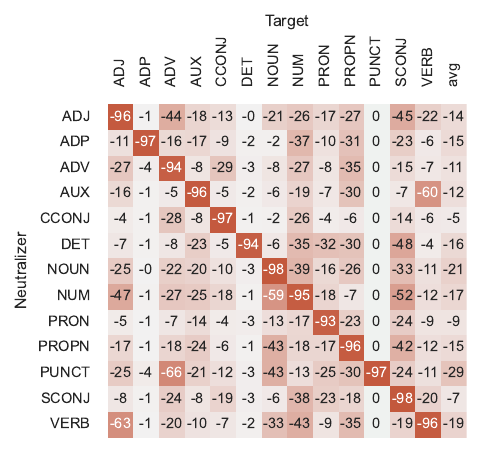}
	\includegraphics[width=0.49\textwidth]{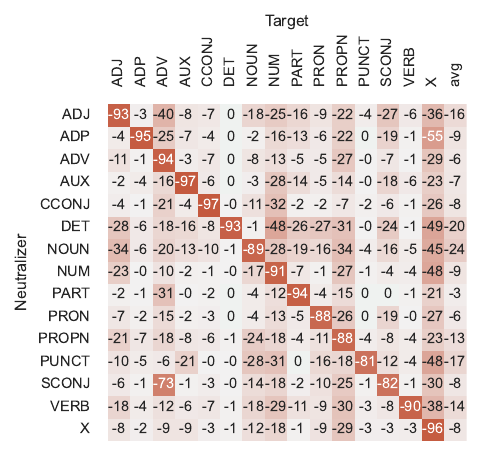}
	\caption{Relative change in accuracy when cross-neutralizing POS tags using centroids from XLM-R.
		In English (top, left), Italian (top right), and Greek (bottom).}
	\label{fig:xneutr_xlm_pos_en_complete}
\end{figure*}

\begin{figure*}[ht]
	\centering
	\includegraphics[width=0.49\textwidth]{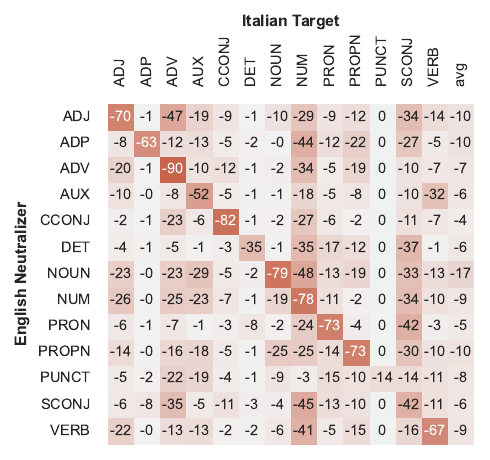}
	\includegraphics[width=0.49\textwidth]{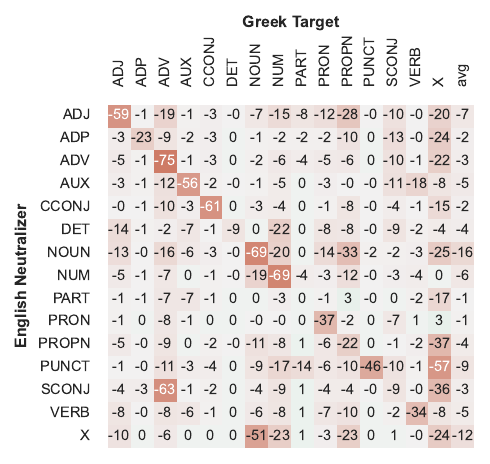}
	\caption{Relative change in accuracy when cross-neutralizing Italian (left) and Greek (right) POS
		tag embeddings using English centroids, using XLM-R representations.}
	\label{fig:xlingual_xneutr_pos_from_en_gum}
\end{figure*}

\begin{figure*}[ht]
	\centering
	\includegraphics[width=0.49\textwidth]{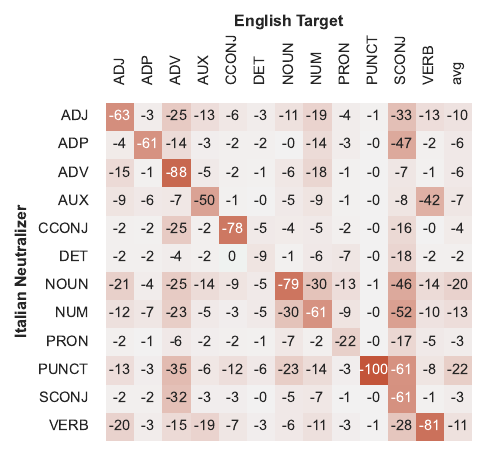}
	\includegraphics[width=0.49\textwidth]{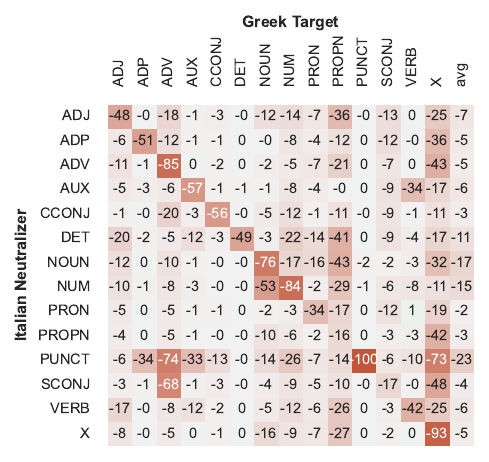}
	\caption{Relative change in accuracy when cross-neutralizing English (left) and Greek (right) POS
		tag embeddings using Italian centroids, using XLM-R representations.}
	\label{fig:xlingual_xneutr_pos_from_it_vit}
\end{figure*}

\begin{figure*}[ht]
	\centering
	\includegraphics[width=0.49\textwidth]{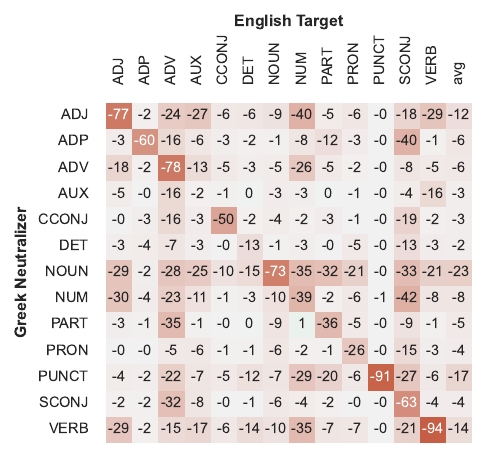}
	\includegraphics[width=0.49\textwidth]{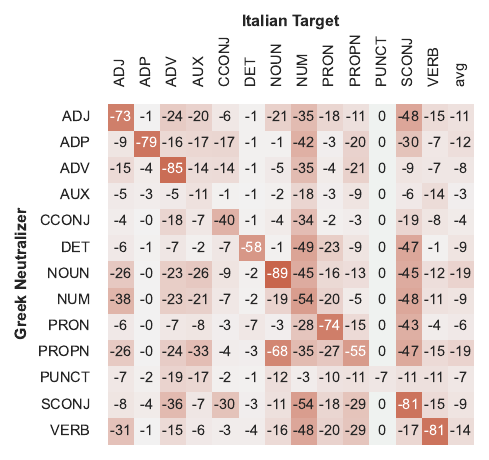}
	\caption{Relative change in accuracy when cross-neutralizing English (left) and Italian (right)
		POS tag embeddings using Greek centroids, using XLM-R representations.}
	\label{fig:xlingual_xneutr_pos_from_el_gdt}
\end{figure*}

\begin{figure*}[ht]
	\centering
	\includegraphics[width=0.8\textwidth]{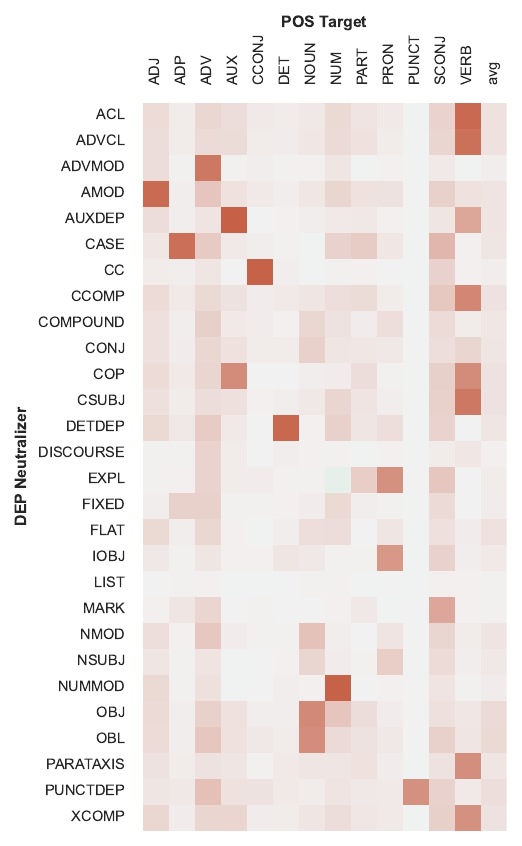}
	\caption{Relative change in accuracy when cross-neutralizing RoBERTa POS embeddings using RoBERTa
		DEP centroids.}
	\label{fig:pos-xt-xn-depneutr}
\end{figure*}

\begin{figure*}[ht]
	\centering
	\includegraphics[width=\textwidth]{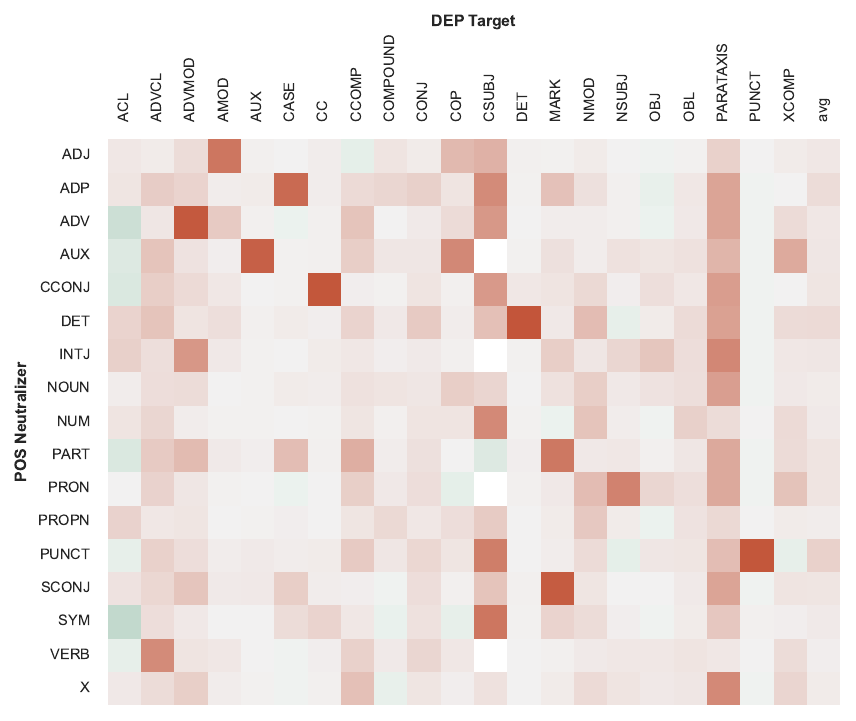}
	\caption{Relative change in accuracy when cross-neutralizing RoBERTa DEP embeddings using RoBERTa
		POS centroids.}
	\label{fig:dep-xt-xn-posneutr}
\end{figure*}

\begin{sidewaysfigure*}[ht]
	\centering
	\includegraphics[width=0.9\textheight]{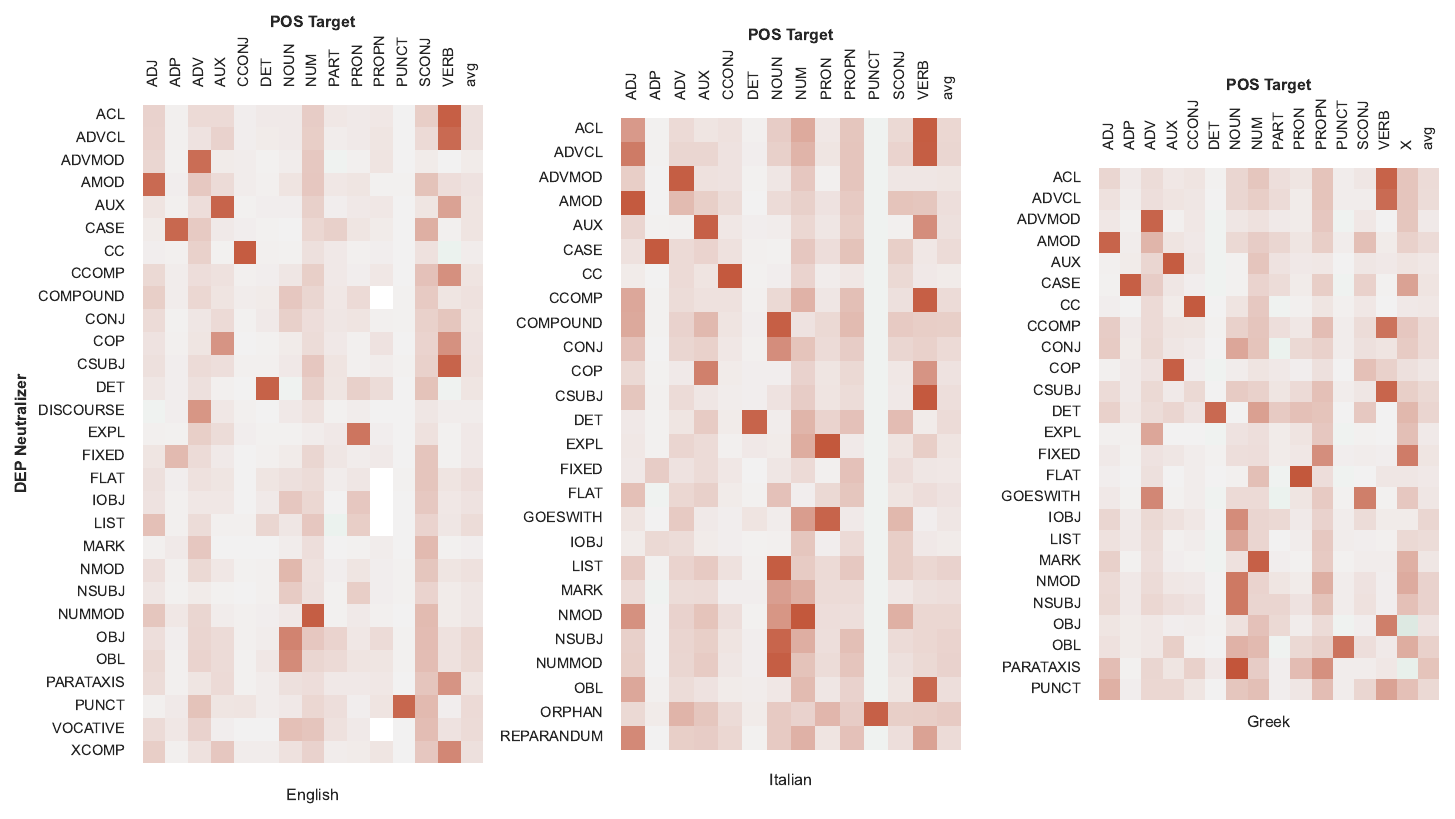}
	\caption{Relative change in accuracy when cross-neutralizing XLM-R POS embeddings using XLM-R DEP
		centroids. In English, Italian and Greek. Some languages have less columns or rows than others
		because the particular category for the missing row/column is not in their dataset.}
	\label{fig:POS-xt-xn-xlm_r-appendix}
\end{sidewaysfigure*}

\begin{figure*}[ht]
	\centering
	\includegraphics[height=0.9\textheight]{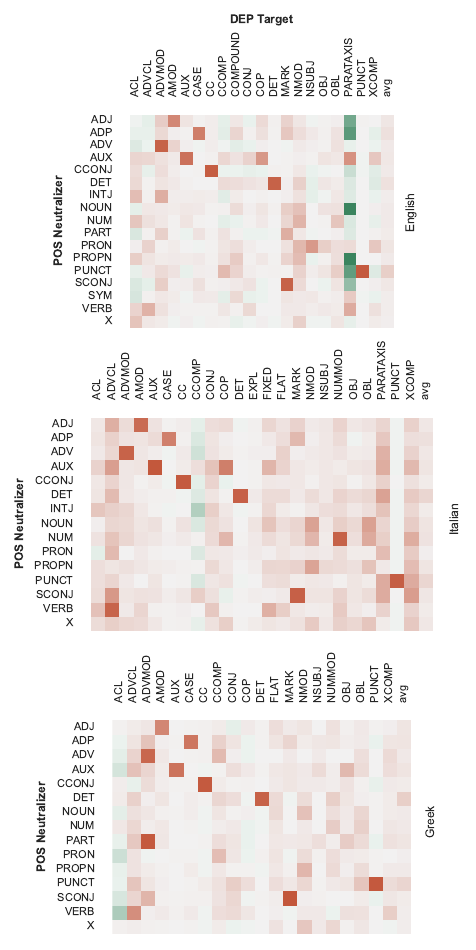}
	\caption{Relative change in accuracy when cross-neutralizing XLM-R DEP embeddings using XLM-R POS
		centroids. In English, Italian and Greek. Some languages have less columns or rows than others
		because the particular category for the missing row/column is not in their dataset.}
	\label{fig:DEP-xt-xn-xlm_r-appendix}
\end{figure*}

We display the complete results for our cross-neutralization experiments in Figs.
\ref{fig:xneutr_roberta_pos_complete} -- \ref{fig:DEP-xt-xn-xlm_r-appendix}. More specifically,
Fig.~\ref{fig:xneutr_roberta_pos_complete}  corresponds to the monolingual setting with RoBERTa on
English, and Fig. \ref{fig:xneutr_xlm_pos_en_complete} to the multilingual setting with XLM-R on
English, Italian and Greek. Figs. \ref{fig:xlingual_xneutr_pos_from_en_gum} --
\ref{fig:xlingual_xneutr_pos_from_el_gdt} show the cross-lingual setting for every possible pairing
of our three languages with XLM-R. Finally Figs.~\ref{fig:pos-xt-xn-depneutr}
and~\ref{fig:dep-xt-xn-posneutr} show the monolingual cross-task setting in RoBERTa in POS-DEP and
DEP-POS directions while Figs.~\ref{fig:POS-xt-xn-xlm_r-appendix}
and~\ref{fig:DEP-xt-xn-xlm_r-appendix} show the same but in the multilingual setting with XLM-R.

\begin{table}[htb]
	\centering
	\footnotesize
	\begin{tabular}{@{}lccc@{}}
		\toprule
		                       & \textbf{Train} & \textbf{Validation} & \textbf{Test} \\ \midrule
		\textbf{English (GUM)} & 4287           & 784                 & 890           \\
		\textbf{Italian (VIT)} & 8277           & 743                 & 1067          \\
		\textbf{Greek (GDT)}   & 1662           & 403                 & 456           \\ \bottomrule
	\end{tabular}
	\caption{Sentence count in each split of each of the datasets we consider, averaging 24 words per
		sentence. }
	\label{tab:datasets}
\end{table}

\begin{table}[htb]
	\centering
	\caption{Cosine similarities between golden and predicted centroids for the RoBERTa POS tagging
		setup.}
	\label{tab:golden_centroids}
	\begin{tabular}{@{}cc@{}}
		\toprule
		\textbf{POS Tag} & \textbf{Cosine Similarity} \\ \midrule
		NOUN             & 1.000                      \\
		ADP              & 1.000                      \\
		ADJ              & 1.000                      \\
		PUNCT            & 1.000                      \\
		AUX              & 1.000                      \\
		VERB             & 1.000                      \\
		DET              & 1.000                      \\
		CCONJ            & 1.000                      \\
		NUM              & 1.000                      \\
		ADV              & 0.9999                     \\
		SCONJ            & 0.9998                     \\
		PRON             & 1.000                      \\
		PART             & 1.000                      \\
		SYM              & 0.9951                     \\
		PROPN            & 0.9999                     \\
		X                & 0.9737                     \\
		INTJ             & 0.9980                     \\ \bottomrule
	\end{tabular}
\end{table}




\end{document}